# A semi-automatic semantic method for mapping SNOMED CT concepts to VCM icons

**Jean-Baptiste Lamy[a], Rosy Tsopra[a], Alain Venot[a], Catherine Duclos[a]**

[a] *LIM&BIO (Laboratoire d'Informatique Médicale et Bioinformatique), UFR SMBH, University Paris 13, Sorbonne Paris Cité, Bobigny, France*

**Abstract**

*VCM (Visualization of Concept in Medicine) is an iconic language for representing key medical concepts by icons. However, the use of this language with reference terminologies, such as SNOMED CT, will require the mapping of its icons to the terms of these terminologies. Here, we present and evaluate a semi-automatic semantic method for the mapping of SNOMED CT concepts to VCM icons.*

*Both SNOMED CT and VCM are compositional in nature; SNOMED CT is expressed in description logic and VCM semantics are formalized in an OWL ontology. The proposed method involves the manual mapping of a limited number of underlying concepts from the VCM ontology, followed by automatic generation of the rest of the mapping.*

*We applied this method to the clinical findings of the SNOMED CT CORE subset, and 100 randomly-selected mappings were evaluated by three experts. The results obtained were promising, with 82 of the SNOMED CT concepts correctly linked to VCM icons according to the experts. Most of the errors were easy to fix.*

## Introduction

Standard medical reference terminologies, such as SNOMED CT, are highly useful for semantic interoperability between health information systems, making it possible to connect electronic health records to decision support systems, epidemiological monitoring systems, etc. However, these terminologies are complex and difficult for clinicians to use [1]. We have therefore developed VCM (Visualization of Concepts in Medicine), a compositional iconic language that represents the main concepts of medical terminologies through icons [2]. VCM is not designed to provide the same level of detail as a textual language. Instead, it provides the clinician with a broader overview, in the form of a graphical summary (*e.g.* from a list of patient problems), or helps the clinician to find the desired items (*e.g.* in a list of search results, after searching in a terminology browser). However, before it can be used with reference terminologies, VCM icons must be mapped to the terms of these terminologies.

H. Saitwal *et al.* [3] considered three methods for mapping medical terminologies: (1) linking together several existing mappings, *e.g.* combining the mappings from SNOMED CT to UMLS (Unified Medical Language System), and from UMLS to ICD10 (International Disease Classification release 10), to create a mapping from SNOMED CT to ICD10 (2) using lexical methods to search for identical or similar terms, and (3) manual mapping, possibly with the use of a specific tool, such as the one proposed by K. Giannangelo et al. to assist the experts for mapping SNOMED CT to ICD 10 [4]. H. Saitwal *et al.* [3] also discussed a fourth method, which they did not test, based on semantic ontology alignment or matching methods [5]; this approach can be used only when the two mapped terminologies are described in description logic (DL), and has therefore rarely been applied to medical terminologies. However, both SNOMED CT and VCM are described in DL, and lexical methods cannot be applied to VCM, due to the lack of textual terms.

Y.R. Jean-Mary *et al.* [6] proposed an algorithm combining lexical and structural methods with semantic verification for mapping between two ontologies. S. Zhang *et al.* [7] aligned several ontologies relating to anatomy. They generated a first set of mappings using lexical methods, and then used the already mapped concepts as "anchors" for mapping other concepts: two concepts from different ontologies having the same relationship with a given "anchor" concept are likely to be equivalent. A similar method was proposed by C. Bousquet *et al.* [8] for mapping the French CCAM (*Classification Commune des Actes Médicaux*, Common Classification of Medical Procedures) to UMLS. In CCAM, terms are defined by up to three descriptors of the anatomical sites involved, the action performed and the mode of access. These descriptors were mapped to UMLS by lexical methods, making it possible to use them as "anchors" for mapping the CCAM procedure terms, which are much more numerous than the descriptors.

The objective of the work presented here was to design a semi-automatic semantic method for mapping SNOMED CT concepts to VCM icons, and to evaluate this method through preliminary mapping for clinical findings. We briefly describe SNOMED CT and VCM, and propose a method for mapping SNOMED CT concepts to VCM icons. We then present the results obtained by applying this method to the SNOMED CT CORE subset, and the results of the review of 100 mappings by three experts. Finally, we discuss the advantages of the method.

## Materials and Methods

### Material

**SNOMED CT** (Systematized Nomenclature of Medicine - Clinical Terms, information about SNOMED CT is available from http://www.snomed.org) is a medical terminology covering various medical concepts, including anatomy, clinical findings and disorders, procedures, organisms, social contexts, etc. SNOMED CT includes many relationships between concepts, including "is a" relationships (*e.g.* hepatitis "is a" hepatic disorder), relationship between clinical findings and finding sites (*e.g.* the liver for hepatitis), associated morphologies (*e.g.* inflammation for hepatitis), and so on. These relationships can be organized into groups (*e.g.* to indicate that the inflammation is located in the liver), particularly for concepts with several finding sites or morphologies. We used the 2012 release of SNOMED CT provided by the National Library of Medicine.

**SNOMED CT CORE Problem List** (Clinical Observations Recording and Encoding) is a subset of 6,286 SNOMED CT concepts. The CORE Problem List was selected to serve as an an appropriate subset for coding clinical information such as the list of patient problems (*e.g.* for discharge diagnosis or reason of encounter). In this study, we used the 2012 version of the CORE Problem List and focused on the 5,345 concepts of this list concerning disorders and clinical findings.

**VCM** [9] is an iconic language for representing the main clinical conditions of the patients. It includes representations of symptoms, diseases, physiological states (*e.g.* age class or pregnancy), risks and history of diseases, drug and non-drug treatments, laboratory tests and follow-up procedures. VCM includes a set of graphical primitives (colors, shapes, and pictograms), and a graphical grammar for combining these elements to create icons.

For the representation of clinical signs and disorders, the focus of this study, a VCM icon can be described in terms of its color, basic shape, set of shape modifiers and central pictogram. The color indicates the temporal aspect of the icon: red for current states of the patient, orange for a risk of future states, and brown for past states. The basic shape is a circle for physiological states or a square for pathological states (diseases or symptoms). The central pictogram indicates the anatomico-functional location (*e.g.* endocrine system) or the patient characteristic (*e.g.* pregnancy) involved; and special pictograms are available for a few specific disorders associated with a specific anatomico-functional location (*e.g.* diabetes for endocrine system). Shapes modifiers can be added to specify (a) a general pathological processes (*e.g.* inflammation or tumor), and (b) a "transverse" anatomical structure that may be present at many anatomico-functional locations (*e.g.* blood vessels, which are present in most organs).

The **VCM ontology** was been designed to formalize the semantics of VCM icons [10]. It has three parts: (1) graphical concepts corresponding to VCM graphical primitives (*i.e.* the various shapes, colors and pictograms), (2) medical concepts (*i.e.* the main anatomical structures, biological functions, pathological processes, *e.g.* liver, hepatic function and inflammation, but *not* the various disorders, such as hepatitis), and (3) relationships between the graphical and medical concepts (*e.g.* the "liver" central pictogram is associated with both the liver (anatomic structure) and hepatic biological function).

**Method for mapping SNOMED CT concepts to VCM icons**

The proposed method for mapping SNOMED CT concepts to VCM icons is based on the compositional nature of both SNOMED CT and VCM. It has two parts: (1) the manual mapping of SNOMED CT concepts to the medical concepts of VCM ontology (n=370), and (2) automatic generation of a mapping of SNOMED CT clinical finding concepts (n=98,590) to VCM icons, using the concepts mapped at step 1 as "anchors" [7], and making use of the relationships present in SNOMED CT and the VCM ontology.

We first generated a manual mapping of SNOMED CT concepts to the medical concepts in the VCM ontology. These medical concepts include the main anatomical structures, biological functions and pathological processes (e.g. liver, hepatic function and inflammation), but not the various disorders (such as hepatitis).

Due to multiple inheritance, some anatomical structures are classified in several branches of SNOMED CT. For example, ear ossicles are classified as both (a) a bone and (b) a part of the ear. The "bone" and "ear" concepts of SNOMED CT respectively map to the "bone" and "ear" concepts of the VCM ontology, which are themselves related to the "bone" and "ear" pictograms. As a VCM icon has only a single central pictogram, ear ossicle disorders would be represented by two icons: one with the "bone" pictogram and the other with the "ear" pictogram. This is problematic because VCM cannot represent the relationship between the two icons, and this may lead the clinician to think that there are two unrelated disorders. Furthermore, clinicians are more likely to associate ear ossicle disorders with the "ear" pictogram than the "bone" pictogram, because they learn about these disorders with auricular disorders in the ENT (Ear-Nose-Throat) specialty and not with bone disorders in rheumatology, as medical education is organized by medical specialty and each disorder is studied in only one specialty. We therefore considered no more than one pictogram per anatomical location and, thus, only disorders affecting several locations were represented by several icons (*e.g.* viral pharyngoconjunctivitis can be represented by an icon for the pharynx and an icon for the eye, because these two sites are different locations).

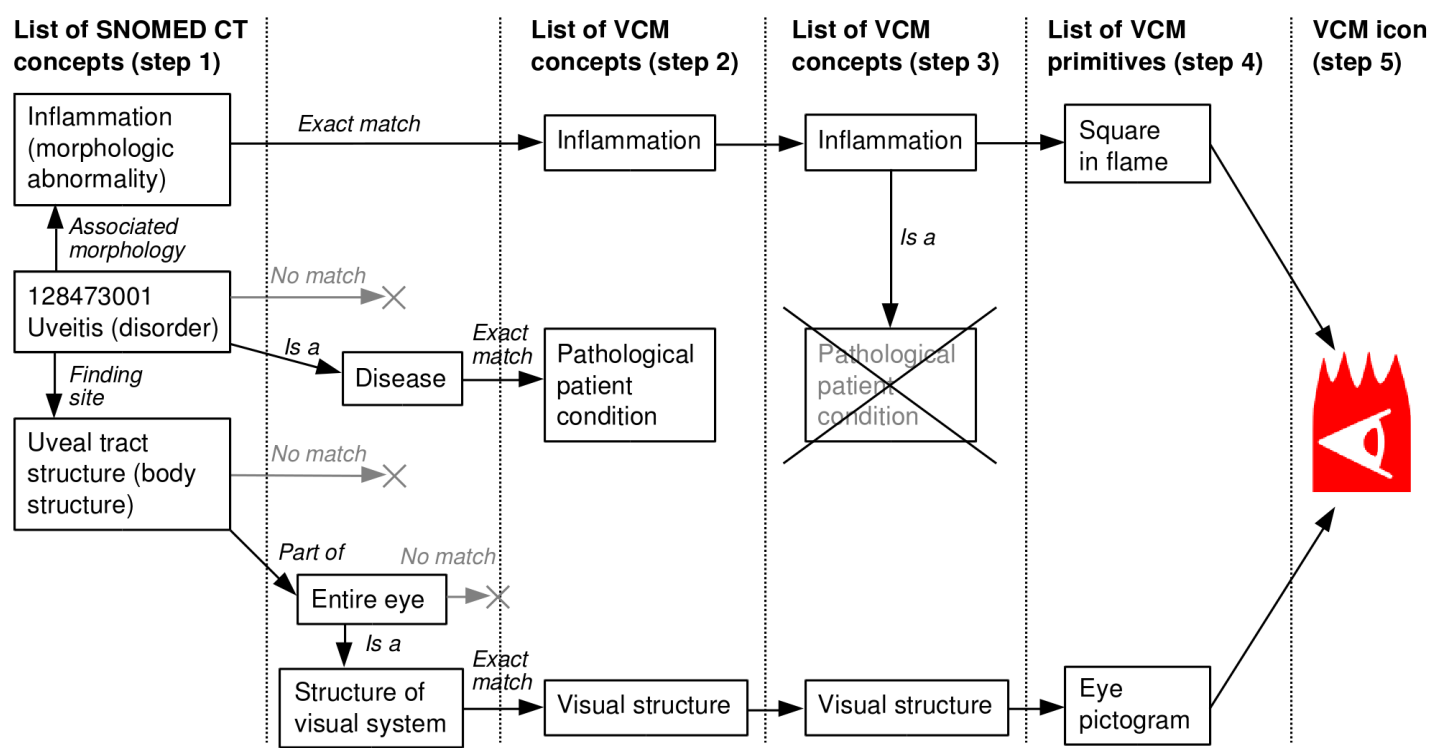

*Figure 1: Example of the method used to associate VCM icons with a SNOMED CT clinical finding concept, applied to the concept of "uveitis". The five steps correspond to those described in the text.*

A preliminary mapping was produced, by considering all the medical concepts in the VCM ontology, and manually searching for the corresponding concepts in SNOMED CT, with a SNOMED CT browser. Each match was marked as either an exact match (*i.e.* the two concepts are equivalent) or a partial match (*i.e.* the SNOMED CT concept is narrower than the VCM concept, and thus not entirely represented by it). We then used a Python script to identify automatically all the anatomical structures in SNOMED CT associated with more than one VCM central pictogram through the ontology. For each of these anatomical structures, a new concept was added to the VCM ontology, and this concept was associated manually with a single VCM pictogram. The choice of pictogram was based on: (a) the medical specialty usually associated to the anatomical structure, (b) the position of the disorders of the anatomical structure, in monoaxial terminologies such as ICD10 (International Classification of Disease, release 10).

We then designed a method for associating one or more VCM icons to a SNOMED CT clinical finding concept. The method consisted in the following steps (see figure 1):

1. Creation of a list of SNOMED CT concepts, including the SNOMED CT concept for which we were generating icons, and each concept related to that concept by one of the following relationships: finding_site, associated_morphology, temporal_context, has_interpretation, interprets, has_definitional_manifestation, pathological_process, has_focus, causative_agent, associated_with, due_to.

2. Creation of a list of concepts from the medical part of the VCM ontology, by mapping each SNOMED CT concept in the list obtained in step 1 to the corresponding concepts (if any) in the manual mapping produced in the first part of the work. For each concept, if an exact match was found, we stopped there and moved on to the next concept. If a partial match or no match was found, we applied the same mapping process recursively to the concept's parents and "bigger-than" concepts (inverse part-of relationship, for anatomical structure: *e.g.* Entire heart for Cardiac valve structure).

3. Retention of only the most specific concepts from the list obtained in step 2.

4. Creation of a list of VCM primitives, by mapping each medical concept in the list obtained in step 3 to the corresponding VCM primitives, using the mapping present in the VCM ontology.

5. Assembly of the VCM primitives obtained in step 4 into one or more VCM icons, as follows: (a) extraction of the list of central pictograms from the list of VCM primitives, (b) extraction of shape modifiers from the list of VCM primitives, and creation of a list of all possible subsets of these shape modifiers, (c) creation of the list of all possible icons, containing each possible combination of a central pictogram (from the list in step 5-a) and a set of shape modifiers (from the set in step 5-b), (d) removal of inconsistent icons, as determined by the VCM ontology, (e) removal of all icons for which there is a more specific icon in the list (*e.g.* if the list contains icons for "hepatic disorder" and "hepatitis", the "hepatic disorder" icon is removed because it is less specific that the "hepatitis" icon).

For SNOMED CT concepts described by several groups, each group was treated separately: we generated, in step 1, a list of SNOMED CT concepts for each group, including the relationships in the group and all relationships belonging to no group (group id 0). These lists were treated as described above, and the resulting icon sets were merged.

**Methods for evaluating the mapping**

For evaluation of the mapping of SNOMED CT concepts to VCM icons, we randomly selected 100 concepts from the clinical findings and disorders concepts in the SNOMED CT CORE Problem List. We generated VCM icons for each of these concepts, and three experts (AV, CD, RT) independently reviewed the icons associated with each concept. Experts were researchers in the field of medical informatics with a medical background (MD or PharmD). During this review, the relationships in SNOMED CT were considered to be the "gold standard", *i.e.* if a piece of information is missing or erroneous in SNOMED CT, it should be expected to be missing or erroneous in the icons. These relationships were extracted from SNOMED CT and made available to the experts.

For each concept, experts had to indicate: (a) whether the icons were acceptable, and (b) any additional comments.

Finally, disagreements between experts were resolved by seeking a consensus by collective discussion.

## Results

**SNOMED CT CORE problem list mapping to VCM**

All 5,345 concepts of the SNOMED CT CORE Problem List corresponding to disorders and clinical findings were associated with VCM icons. In total, 4,874 concepts (91.2%) were associated with a single icon, 435 concepts (8.1%) with 2 icons, 32 concepts (0.6%) with 3 icons, 3 concepts with 4 icons and 1 with 5 icons. There were 758 different VCM icons in the mapping, so each icon corresponded to a mean of 6.8 concepts. Table 1 shows examples of SNOMED CT concepts and the corresponding VCM icons.

Only 327 concepts (6.1%) from the CORE Problem List were associated with a VCM icon without a central pictogram and shape modifier (*i.e.* a very general icon indicating nothing more than "disorder"). These SNOMED CT concepts included mostly clinical findings (rather than disorders). Most were either (a) loosely defined clinical signs, *e.g.* "General symptom" (267022002), or (b) related to drug prescriptions, medical procedures or laboratory tests, *e.g.* "Already on aspirin" (405748007), "Transplant follow-up" (183655000), and "Lithium monitoring" (275917000). Although these concepts are classified as clinical findings in SNOMED CT, they fall beyond the scope of this study, which focused on clinical signs and disorders.

The manual mapping between the medical concepts of the VCM ontology and SNOMED CT involved 334 concepts from the ontology and 1,752 SNOMED CT concepts, a ratio of 5.2. During the design of the mapping process, 181 SNOMED CT anatomical structures were initially associated with more than one VCM central pictogram. For these concepts, 97 new concepts were added to the VCM ontology, and these concepts were then manually associated with a single central pictogram.

**Evaluation results**

The icons associated with 100 randomly selected SNOMED CT concepts were considered, by the three experts, to be acceptable for 82 icons, and the experts considered three of the concepts to be beyond the scope of the study (*i.e* procedures, laboratory tests or treatments, rather than clinical findings).

The erroneous icons generated for the 15 remaining concepts were analyzed and classified. Most related to missing associations in the manual mapping between the VCM ontology and SNOMED CT (8 errors, *e.g.* disorder of stature was not associated with the size concept in VCM ontology; it icons therefore did not include the size pictogram) or erroneous associations (2 errors, *e.g.* complication of procedure were wrongly associated with iatrogenic disorders). Two errors were related to the choice of medical specialty associated with a given anatomical structure (*e.g.* maxillary bone disorders are associated with the ENT specialty, and should therefore not be associated with the bone pictogram). Three errors were related to an absence of information in SNOMED CT for the pathological or abnormal status (*e.g.* for the Lung field abnormal (274710003) concept, no relationship in SNOMED CT made it possible to deduce that the concept was abnormal), and for hypo/hyperfunctioning.

## Discussion and conclusion

We present here a semi-automatic method for mapping SNOMED CT concepts to VCM icons. This method involved manual mapping for a limited number of concepts (370 concepts from the VCM ontology) followed by automatic mapping to any SNOMED CT concept. The method was successfully applied to, and evaluated on, the clinical finding concepts from the SNOMED CT CORE subset.

The compositional structures of VCM and SNOMED CT are very similar, even though VCM was not based on SNOMED CT. Both describe disorders in terms of sites, morphologies and etiologies. Two main differences were encountered during manual mapping between the concepts of the VCM ontology and SNOMED CT: (1) VCM distinguishes pathological / abnormal patient conditions, *versus* physiological conditions, whereas, in SNOMED CT, the distinction is between disorders and clinical findings, and there is no concept for abnormal non disorder clinical findings; and (2) SNOMED CT has no concepts for describing disorders of biological functions, *e.g.* there is no "hyperfunction" concept associated with the "hyperthyroidism" concept (see table 1). These differences resulted in extra work during manual mapping and searching in the SNOMED CT browser: we searched for "abnormal" and various synonyms, and when we mapped a biological function, we searched for the function itself and for the associated hypo- and hyperfunction, *e.g.* for "thyroid function" we also searched for "hypothyroidism" and "hyperthyroidism".

A few more specific differences were encountered between VCM and SNOMED CT: (a) VCM represents pain as a symptom, with a specific shape modifier, whereas, in SNOMED CT, pain is considered as a sensory nervous system finding, (b) VCM considers the autoimmune aspect of a disorder as an etiology, whereas SNOMED CT considers autoimmune disorders as immune system disorders, (c) chromosomal abnormalities are considered as genetic diseases in VCM, whereas SNOMED CT classifies them as congenital malformations, (d) VCM considers muscles and bones to be separate structures, whereas SNOMED CT considers some skeletal muscles to be part of the skeletal system, *e.g.* the structure of the psoas major muscle (id 57240007) is (recursively) a part of the skeleton system structure (id 113192009); this might be a problem in SNOMED CT.

*Table 1 - Examples of SNOMED CT concepts and their associated VCM icons*

| SNOMED CT concept | Principal SNOMED CT relationships | VCM Icon | Meaning of the icon |
|---|---|---|---|
| 34486009 Hyperthyroidism (disorder) | finding site: Thyroid structure<br>is a: Disease | 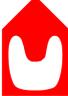 | Hyperthyroidism |
| 36184004 Aneurysm of renal artery (disorder) | associated morphology: Aneurysm<br>finding site: Structure of renal artery<br>is a: Disease | 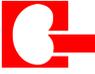 | Renal vascular disorder |
| 25416002 Peripheral neuralgia (disorder) | has definitional manifestation: Pain<br>finding site: Peripheral nervous system structure<br>is a: Disease | 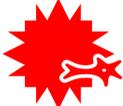 | Peripheral nerve pain |
| 4927003 Acute anterior uveitis (disorder) | associated morphology: Acute inflammation<br>finding site: Anterior uveal tract structure<br>is a: Disease | 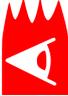 | Ocular inflammation |
| 254937005 Intracranial glioma (disorder) | associated morphology: Glioma<br>finding site: Brain structure<br>is a: Disease | 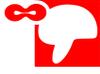 | Tumor in the central nervous system |
| 186675001 Viral pharyngoconjunctivitis (disorder) | causative agent: Virus<br>finding site: Conjunctival structure<br>finding site: Pharyngeal structure<br>is a: Disease | 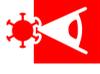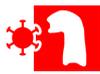 | Ocular viral infection<br>ENT viral infection |

We chose to restrict the mapping to one icon per anatomical location affected by the clinical finding, to limit the number of icons used and to stick to the medical specialties usually associated with particular conditions. This choice is debatable, but the importance of medical specialty for data presentation to clinicians has been highlighted elsewhere, by P. Massari *et al.* [11]. Relationships between a disorder and anatomical structures outside of the disorder's medical specialty (*e.g.* the relationship between ear ossicle disorders and bones) are ontologically true and useful for reasoning in a reference terminology, such as SNOMED CT, but they are not necessarily appropriate for presentation to clinicians in an interface terminology [12] or an iconic language, such as VCM.

The semi-automatic method we propose for mapping SNOMED CT concepts to VCM icons has several advantages over manual methods. Previous attempts to map VCM icons and medical terminologies manually have been based on MeSH, ICD10 and ATC. However, in manual validations of these mappings, the level of inter-expert agreement was low. Most of the problems encountered by the experts were not directly related to VCM, instead relating to the definition of disorders (*e.g.* "what are the anatomical structures affected by this disorder?"). In our semi-automatic method, many of these problems were solved automatically, by considering the relationships in SNOMED CT. In addition, when a new version of VCM or SNOMED CT is available, it should be easier to update the mapping if this can be done semi-automatically.

Our method is similar to the "anchor"-based semantic mapping methods proposed by S. Zhang *et al.* [7]. Semantic methods were possible given the ontological nature of both SNOMED CT and VCM. These semantic methods should be more accurate than lexical methods, because DL descriptions of concepts are more expressive than textual labels [13], *e.g.* in textual label, "and" sometimes means a logical AND but frequently means a logical OR, such as in the term "structural and functional abnormalities of the kidney".

The evaluation of 100 random mappings yielded promising results. Most of the errors encountered were related to elements missing from the manual mapping. These errors were not serious as they generated icons that were too generic, but nevertheless appropriate (*e.g.* the icon for cardiac disorders instead of the icon for heart rhythm disorders), and they were easy to fix by complementing the manual mapping. The evaluation we presented was limited to clinical findings, however it would also be interesting to evaluate the coverage of the various SNOMED CT axes (such as anatomy or morphology) by the corresponding pictograms in VCM.

In conclusion, we present here a semi-automatic method for mapping the concepts of SNOMED CT, a reference terminology, to the VCM icons. Future perspectives of this work include: (a) the mapping of the laboratory tests, procedures or drug treatment concepts of SNOMED CT to VCM icons by the same method, (b) an analysis of the expressiveness of VCM with respect to SNOMED CT and, possibly, the extension of VCM to improve coverage, (c) mapping of terminologies to VCM, *e.g.* VANDF-RT (Veteran Administration's National Drug File Reference Terminology) for drugs, either by a similar method or by combining the SNOMED CT to VCM mapping with the existent mappings in UMLS, and (d) the use of VCM icons to display elements of electronic patient records coded in SNOMED CT.

## Acknowledgments

This work was partly supported by the French National Research Agency (ANR) through the TecSan program (project L3IM n°ANR-08-TECS-007).

**Address for correspondence**

Jean-Baptiste Lamy <jibalamy@free.fr>, Bureau 149, UFR SMBH, 74 rue Marcel Cachin, 93017 Bobigny cedex, France